\documentclass[conference]{IEEEtran}
\IEEEoverridecommandlockouts
\usepackage{cite}
\usepackage{amsmath,amssymb,amsfonts}
\usepackage{graphicx}
\usepackage{textcomp}
\usepackage{xcolor}
\usepackage{balance}
\usepackage{wrapfig, blindtext}
\usepackage{algpseudocode}	 
\usepackage[hmargin=.75in,vmargin=1in]{geometry}
\usepackage[american]{babel}
\usepackage[T1]{fontenc}
\usepackage{times}
\usepackage{comment}

\usepackage{tikz}
\begin{document}

\title{Evolutionary Algorithms and Efficient Data Analytics for Image Processing
}

\author{

\IEEEauthorblockN{Farid Ghareh Mohammadi$^{1}$, Farzan Shenavarmasouleh$^{1}$, M. Hadi Amini$^{2}$, and Hamid R. Arabnia$^{1}$}
\IEEEauthorblockA{\textit{$^{1}$ Department of Computer Science, University of Georgia, Athens, GA, USA} \\
\textit{$^{2}$ School of Computing \& Information Sciences, Florida International University, Miami, USA}\\
\{farid.ghm,fs04199,hra\}@uga.edu, moamini@fiu.edu}

}

\maketitle

\begin{abstract}
Steganography algorithms facilitate communication between a source and a destination in a secret manner. This is done by embedding messages/text/data into images without impacting the appearance of the resultant images/videos. Steganalysis is the science of determining if an image has secret messages embedded/hidden in it. Because there are numerous steganography algorithms, and since each one of them requires a different type of steganalysis, the steganalysis process is extremely challenging. Thus, researchers aim to develop one universal steganalysis to detect all known and unknown steganography algorithms, ideally in real-time. Universal steganalysis extracts a large number of features to distinguish stego images from cover images. However, the increase in features leads to the problem of the curse of dimensionality (CoD), which is considered to be an NP-hard problem. This COD problem  additionally makes real-time steganalysis hard. A large number of features generates large datasets for which machine learning cannot generate an optimal model. Generating a machine learning based model also takes a long time which makes real-time processing appear impossible in any optimization for time-intensive fields such as visual computing. Possible solutions for CoD are deep learning and evolutionary algorithms that overcome the machine learning limitations. In this study, we investigate previously developed evolutionary algorithms for boosting real-time image processing and argue that they provide the most promising solutions for the CoD problem.
\end{abstract}

\begin{IEEEkeywords}
 Image steganalysis, Image classification, Feature extraction, Feature selection, Curse of dimensionality, Dimension reduction, NP-hard problem, Data science
\end{IEEEkeywords}

\section{Introduction}
\noindent $\diamond$ \textbf{Motivation}
 Nowadays, researchers seek a way to distinguish between cover images and stego images focusing on a specific steganographic scheme in a fraction of a second. However, the problem occurs when we do not know the scheme; which makes steganalysis challenging, especially in real-time processing. To solve this problem, scientists utilize a universal steganlaysis which enables them to recognize if a given image is a stego or cover. This universal steganalysis has to detect all steganographic schemes to become able to classify images by extracting important features. The more features we extract, the higher will be the chance of detecting the steganographic schemes that lead us to classify images accurately. However, the high number of features generates a high dimensional problem known as the curse of dimensionality (CoD). This is an important problem to solve because analyzing the features in details requires a lot of time and is not easy to do in real-time. Furthermore, prior works \cite{kouiroukidis2011effects} \cite{ermon2013taming}  have found the CoD problem to be an NP-hard problem and have attempted to solve it by reducing the dimensionality of the data. 
 
\begin{figure}[ht]
    \centering
  \includegraphics[height=2.in]{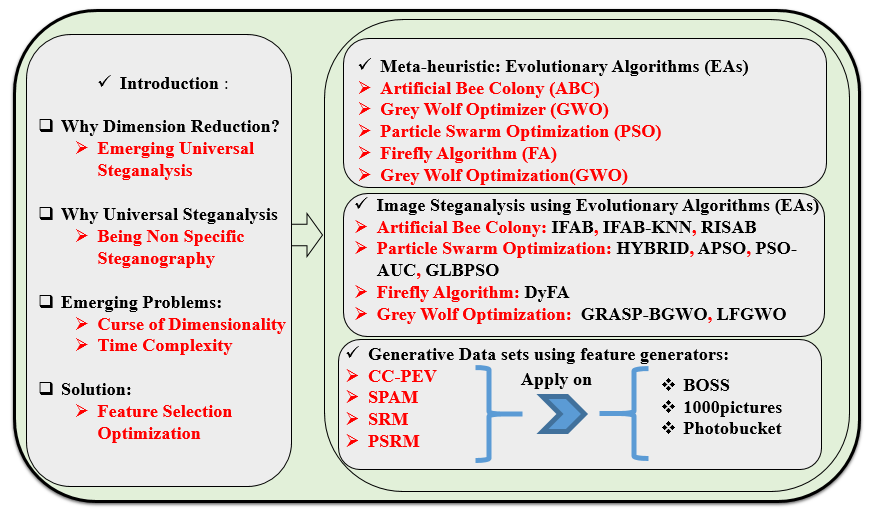}
    \caption{A general overview of this study}
    \label{fig:ISEA_Overview}
\end{figure}

 In this study, we aim to address this problem and present the potential solutions. Figure \ref{fig:ISEA_Overview} illustrates the overall structure of this paper.
 Steganography is an advanced skill and communication method that allows hidden secret messages to be sent via an innocuous multimedia file \cite{boroumand2019deep}. The most common multimedia communications involve images, audios, videos, even text files, or internet protocols \cite{Farid2012}. The cover multimedia, with secret messages embedded in, is called stego multimedia.  The goal of steganography is to make the stego multimedia and the associated cover look as identical as possible. Otherwise, the risk of detection would be high.
Steganalysis is a challenging field because all the different steganography schemes must be recognized using different steganalysis algorithms. Researchers have proposed steganalysis algorithms for individual steganography schemes. To this end, first, they extract important features based on preferred extraction
strategies \cite{kodovsky2009calibrationCCPEV}. The goal of these features is to provide a distinguishing border between stego images and cover images. 
Further, researchers propose a universal steganalysis to detect stego images from cover ones without considering the steganographic scheme, known or unknown, used to embed the hidden messages. A large number of research studies have proposed to improve steganalysis performance with a large number of features like CC-C300 with 48600 features \cite{kodovsky2011steganalysisCC-C300}, in which the the authors aimed to identify the goodness of utilizing high-dimensional features, together with ensemble classifiers by comparing them to selected existing steganalysis algorithms. Another example is PHARM with 12600 features \cite{holub2015phasePHARM}, in which the authors identified an important
discovery that in the pixels of a decompressed JPEG image are not shift invariant and their statistical characteristics heavily rely on their position in the filtering (e.x. 8*8 grid). These are examples of the universal (blind) approach. However, to do universal steganalysis like those mentioned above, we need to extract as many features as possible,  which is also the main disadvantage of it. Because this leads to the problem of the Curse of dimensionality (CoD). CoD is generated upon extracting or gathering too much information.

 \noindent $\diamond$ \textbf{Related Works}   Deep learning tries to enhance the detection accuracy while considering high dimensional images. For example, Boroumand \emph{et al} \cite{boroumand2019deep} proposed a novel convolutional neural network (CNN) architecture called SRNet for steganalysis. SRNet is independent of many newly introduced design elements yet it still produces a high performance through deep learning. Although deep learning enhances the performance of steganalysis significantly, it still has time complexity issues. Along side of deep learning, evolutionary algorithms (EA) have been proposed for this task and has shown that they can improve the performance of steganalysis better than deep learning \cite{back2018evolutionary}. 

It is imperative to develop efficient data analytics algorithms to deal with emerging engineering problems, e.g., critical infrastructures\cite{amini2020interdependent}. Evolutionary algorithms lend themselves as powerful computational algorithms \cite{back2018evolutionary} that can utilize a large number of techniques to solve large-scale engineering and science problems .  EA provides an environment in which steganalysis yields a low time complexity. Researchers have adopted evolutionary algorithms for a wide variety of purposes in steganalysis like feature selection, finding the most probable sub-images in spatial domain, and much more. In this study, we focus on evolutionary algorithms that have attempted to improve steganalysis, such as Artificial Bee Colony (ABC) \cite{karaboga2005idea}.
These studies \cite{ch2_farid} utilized feature selection techniques to solve the COD problem \cite{ch1_farid}.

\section{Image steganalysis using evolutionary algorithms}
\begin{figure}[ht]
    \centering
  \includegraphics[height=2in]{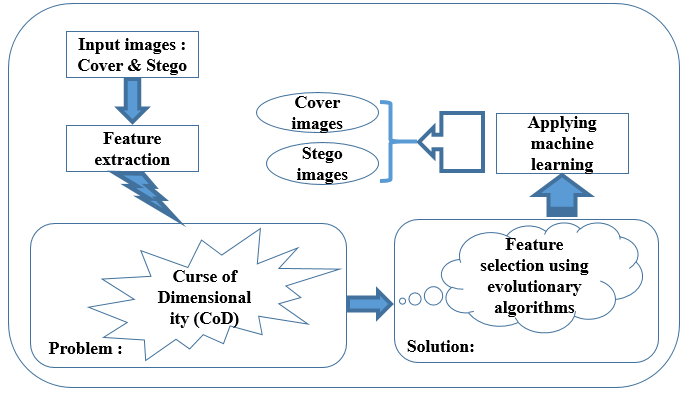}
    \caption{A general flowchart of image steganalysis using evolutionary algorithms \cite{ch1_farid} \cite{ch2_farid} }
    \label{fig:EAsteganalysisGeneral}
\end{figure}

Recently, researchers have sought to optimize the process of steganalysis, particularly within image domain. Through their research they developed a successful algorithms which will be discussed further in detail. The general overview of the combination of evolutionary algorithms and steganalysis are shown in Figure \ref{fig:EAsteganalysisGeneral}.

The goal of universal steganlaysis is to find a model that performs optimally on both known and unknown steganography schemes, however, this leads to  the curse of dimensionality. Researchers have attempted to solve the CoD problem using evolutionary algorithms. EAs aim to minimize the number of feature dimensions as much as possible while maintaining or improving performance. EAs always choose an optimal feature dimension, which will likely yield a significantly higher performance compared to the traditional machine learning feature selection algorithms.     

Scientists deal with the large-scale engineering and science projects that seek optimization either by finding global maximum or global minimum in a continuous or discrete space. However, due to the presence of various local maximums or minimums, traditional machine learning algorithms run at the risk of reaching them instead, and rarely finding global maximum or minimum. The likelihood of being stuck in local maximum or minimum is further exacerbated by CoD. Therefore, researchers leverage EAs to seek global maximum or minimum solutions by skipping local ones. Universal steganalysis, specifically, requires evolutionary algorithms to solve this problem in discrete space. EAs are used as a feature selection tool to decrease the feature dimensions in discrete space to reach the global maximum solution, i.e. the most relevant subsets of features which yield the highest accuracy on detecting stego images from cover images.\par

 \noindent $\diamond$ \textbf{Contributions} 
 The main contribution of this paper is to address the main problem of big data, COD, and determine the most promising solutions and trends to tackle this problem. Scientists have conducted research studies to solve COD and make the machine learning algorithms get trained well and robust enough to yield high performances. So, our contributions are as follows:\par
 $\diamond$ Recognising one of the most promising solutions to COD problem that is feature selection using evolutionary algorithms. We discuss state-of-the-art works that solved this problem with and without evolutionary algorithms.\par
 $\diamond$ Discovering the trends for image classification and image steganlaysis. We aim to find techniques that have used evolutionary algorithms to optimize the process of steganalysis.
 
\section{Image steganalysis using artificial bee colony }
The idea of the artificial bee colony (ABC) algorithm, as developed by Karaboga \cite{karaboga2005idea}, is more suitable for continuous and discrete optimization problems. ABC algorithm is a powerful optimization algorithm, which simulates the foraging behavior of honeybees to find the best food source. ABC works based on three types of bees: 1) employed bees, 2) onlooker bees and 3) one scout bee. ABC starts with employed bees computing the goodness of each solution, then onlooker bees seek the global maximum solution with respect to the best goodness using cross over, which generates new offspring out of selected high solutions to create a better solution. This process repeats itself until no higher maximum goodness is found or the number of crossover iterations meet the pre-designed limit. After that, the scout bee goes beyond of the boundary of onlookers to seek global maximum goodness (solution). 
This entire process repeats until the condition is met or again it reaches the maximum number of iterations. 
Onlooker bees run until the limit condition is satisfied. Next, one of employed bees converts into a scout. Note that, in each iteration, Scout performs mutation, which is the final phase of evolutionary algorithms.

Researchers have proposed a new customized version of ABC to solve their discrete engineering problems such as CoD. There are plenty of customized ABC applications particularly for image processing, including but not limited to  \cite{Farid2014IFAB} \cite{mohammadi2017RISAB}. Owing to its simplified implementation, ABC is a widely-used algorithm with flexibility, robustness and ability to exploit local solutions. It is also capable of exploring the whole sample space to find the global solution. The most important advantages of ABC are its power to solve any problem and its simplicity, which is duo to the minimum number of parameters to be tuned. Therefore, ABC seems to be far better than other evolutionary algorithms like Genetic Algorithm (GA). However, ABC suffers from a low convergence rate in sequential processing and slow execution processing to compute precise solutions. These are the reasons that ABC may not obtain the best solution.\cite{gerhardt2012ABCartificial}

\subsection{Image steganalysis using Feature Selection based on ABC}
Mohammadi and Saniee Abadeh in \cite{Farid2013Conference} first presented the idea of dimension reduction using evolutionary algorithms like ABC, then in \cite{Farid2014IFAB} proposed a new steganalysis method named IFAB, which helps steganalysis to enhance the performance of distinguishing stego images from the cover images. Authors adopted a wrapper-based feature selection technique by customizing the original ABC into the discrete ABC algorithm. ABC selects the feature subsets, and a classifier is employed to evaluate every feature subset generated by the algorithm. According to the IFAB algorithm for feature selection, the process of feature selection solves the problem of CoD by leveraging employed bees, onlooker bees and a scout bee.

According to the table \ref{tab:comparisonResearches}, Mohammadi's and Saniee's study \cite{Farid2013Conference}, and a more comprehensive study \cite{Farid2014IFAB} would be considered the firsts to apply the EAs on steganalysis. IFAB significantly minimizes the feature dimension and selects 80 features out of 686 SPAM (Subtractive Pixel Adjacency Matrix) attributes. IFAB also shrinks the CC-PEV (Cartesian Calibrated features by PEVny) feature dimension well by selecting 250 features out of 548. The Final accuracy of IFAB is 60.98 and 68.22 for SPAM and CC-PEV datasets, respectively. 

\begin{table}[h]
    \centering
    \caption{ Research study Comparison}
    \begin{tabular}{|c|c|c|c|c|}
    \hline
 Paper & Evol. alg. & Proposed method & FS type& Year   \\ [0.5ex] 
 \hline
  \cite{Farid2014IFAB}  & ABC    &IFAB  & Wrapper& 2014\\
   \cite{mohammadi2014IFABKNN} & ABC    & IFAB-KNN&Wrapper &2014\\
   \cite{mohammadi2017RISAB} & ABC  & RISAB &Wrapper  &2017 \\
    \cite{chhikara2016hybridPSO}  & PSO   & HYBRID &Wrapper \&Filter  & 2016\\
   \cite{adeli2018imagePSO} & PSO    &  APSO &Filter&2018\\
   \cite{rostami2016PSO} & PSO    & PSO-AUC  &Filter & 2016 \\
    \cite{kumari2017GLBPSO}  & PSO   &  GLBPSO  & Wrapper& 2017 \\
    \cite{kaur2018feature} & PSO & MI-APSO & Wrapper$^*$&2018\\
   \cite{chhikara2018improvedFA} &   FA  &  DyFA &Wrapper & 2018\\
   \cite{veena2019GRASPBGWO} &  GWO & GRASP-BGWO &Wrapper   &2019\\
   \cite{pathak2019GWO} & GWO & LFGWO &Wrapper &2019 \\
   \cite{Guttikonda2019} &PGO &GLCM-PGO&Wrapper$^*$ &2019\\
   \hline
 
    \end{tabular}

   \footnotesize{ $^*$ goes to special kind of wrapper that select features by eliminating the non-informative features. The rest of papers select subsets of features regularly }
    \label{tab:comparisonResearches}
\end{table}
\subsection{comparative studies}
Mohammadi and Saniee Abadeh in \cite{mohammadi2014IFABKNN} presented an improved evolutionary algorithm approach for image steganalysis to enhance IFAB, named IFAB-KNN. IFAB-KNN also provided a wrapper based feature selection algorithm. The authors embedded K-Nearest Neighbor(KNN) into ABC to help it evaluate each subset of features more carefully. Concretely, KNN plays a significant role as a fitness function for evaluating subset features in ABC. IFAB-KNN outperforms IFAB with the updated tuning parameters while keeping the same number of selected features.

Mohammadi and Saniee Abadeh in \cite{mohammadi2017RISAB} proposed a new hybrid approach to steganalysis named, region based Image Steganalysis using Artificial Bee colony (RISAB). RISAB enables ABC to search in image space, particularly spatial domain, to find the most probable sub-image to carry the hidden messages. As a matter of fact, the likelihood of embedding messages in a sub-image is high if the amount of intensity and energy is greater than the other parts of the image. RISAB is a combination of applying IFAB over the whole image, and the sub-image selected by ABC. The first main step and goal of their paper is to investigate the whole image to identify such a high energy sub-image leveraging a customized ABC to go through the spatial domain of the image. Then, they extract features twice. First, they extract features from the whole given image. Second, they extract the same features from the sub-image found in the earlier phase.  They apply the same features that IFAB  \cite{Farid2014IFAB} presents, which are the best subset of features for feature extractors, SPAM and CC-PEV. Having extracted the features from the whole image and sub-image, researchers merge the datasets for both feature extractors separately. The final dataset for SPAM and CC-PEV involves 160 and 500 features, respectively. In either datasets, features visualize the instances in a way that the classifier will be able to train and find a high-performance model. Since authors considered both the whole images and the associated sub-images, their approach greatly outperforms IFAB. 

In short, Mohammadi and Saniee Abadeh applied ABC on two different areas: Numeric data and Spatial domain. They proposed two unique methods, so called IFAB and RISAB. Both of them greatly increased the performance of steganalysis due to their dimension reduction properties while solving the CoD problem. It is also worth noting that in the state-of-the-art work ABC is being used to detect Malwares. \cite{mohammadi2020malware}

\section{Image steganalysis using particle swarm optimization}
Particle Swarm Optimization (PSO) \cite{kennedy1995eberhartPSO} is an optimization method to solve non-linear optimization problems, presented by Kennedy and Eberhart. PSO is inspired by the behaviour of a flock of birds or fish swarms. Several researchers proposed to extend a version of PSO to solve optimization problems, particularly for steganalysis. \cite{chhikara2016hybridPSO} \cite{adeli2018imagePSO}

Chhikara \emph{et. al} \cite{chhikara2016hybridPSO} proposed a new hybrid approach using PSO for steganalysis named HYBRID. They proposed a hybrid filter and wrapper based feature selection approach to deal with the computational complexity in image steganalysis. The researchers examined their approach for attacking different steganography algorithms.The customized PSO improved the classification accuracy of detecting stego images and cover images, while reducing time complexity of the process as well. The authors enhanced the accuracy of classification significantly for SPAM and CC-PEV up to 10 percent and 14 percent, respectively while examining the HYBRID algorithm on the different steganography algorithms.

From another point of view towards steganalysis, a few researchers presented a novel approach to solve the problem of discovering the message embedded in a covert multimedia using Area Under Curve (AUC) as a new measure for fitness function to evaluate the selected features \cite{rostami2016PSO} \cite{adeli2018imagePSO}. They considered another filter-based feature selection algorithm for steganalysis. The authors \cite{adeli2018imagePSO} introduced an Adaptive inertia weight-based PSO called APSO. APSO is adopted for steganalysis with two main phases. First, there is a feature selection process, which decreases the dimensionality of features; second making a final model to recognize stego images from cover images using the selected subsets of features. APSO used a novel fitness function which provides Area Under Curve (AUC) to evaluate the selected feature subset. To evaluate the given approach, authors used several classifiers such as SVM, DT, NB, and KNN. The SVM obtained the best result in comparison with the others when the hyper-plane experienced the largest distance between support vectors of given stego and cover classes. APSO-AUC decreased the feature dimension and selected top 140 features out of 686 SPAM attributes and 363 features out of 548 features. The Final accuracy of APSO using SVM yields 82.62 and 87.72 for SPAM and CC-PEV datasets, respectively. These methods yield a better result using PSO in comparison with IFAB and RISAB because the time complexity has been reduced on large data sets. 

Chikara and Kumari in \cite{kumari2017GLBPSO} proposed another version of PSO for steganalysis. The authors introduced a new wrapper-based feature selection, named Global Local PSO (GLBPSO). They adopted 
neural networks to evaluate the selected feature subsets by GLBPSO.
The GLBPSO algorithm improved the standard PSO by having the best global and local PSOs simultaneously. In \cite{kumari2017GLBPSO}, researchers take Chen's approach \cite{chen2008jpeg}, and decrease its dimension by selecting the best feature subset. The prediction performance of GLBPSO provides no more than 7 percent improvement in comparison with the basic results, where performance is calculated based on all features. GLBPSO reduced features down to 282 features out of 486 features. Furthermore, Kaur and Singh \cite{kaur2018feature} proposed a new feature selection leveraging mutual information and adaptive PSO (MI-APSO) using area under curve for image steganalysis. MI-APSO was also inspired by IFAB as a feature selection technique and further improved the performance of image steganalysis.

\section{Image steganalysis using firefly algorithm}
Yang introduced a new evolutionary algorithm in \cite{Yang2008Firefly} named Firefly algorithm (FA) which is inspired by the
flashing behaviour of the fireflies. FA aims to attract other objects, namely, their mates, through their light. Yang presented this algorithm based on the following assumption: all fireflies are the same gender. It means that the possibility of attacking other mates is the same. Also the rate of attraction has a direct relation with the light; the more light, the higher the rate of attraction. Furthermore, if there are no bright fireflies, mates tend to move towards any of the fireflies randomly. Researchers developed and tailored FA to work well with steganalysis.

It is worth mentioning that firefly algorithm can improve the convergence rate problem of ABC. In this section we investigate the application of FA to steganalysis.

Chikara \emph{et al}\cite{chhikara2018improvedFA} proposed a new dynamic algorithm to stegalysis, named Dynamic firefly (DyFA). They customized firefly algorithm for universal steganalysis. Feature Selection (FS) is the main role of DyFA, which alleviates the computational complexity of universal steganalysis. FA provides two important parameters alpha and gamma, which help FA to converge faster per each iteration. Basically, tuning these parameters is important for FA to accelerate the process of feature selection. In addition, the researchers apply a hybrid FA as DyFA, which combines the filter method (t-test and regression) and wrapper method for FS. The results reveal that DyFA reduces features by about 77$\ -\ $93 percent of the original feature dimension, which improves the accuracy of distinguishing stego images from cover images by about 2$\ -\ $10 percent. Like other discussed studies, the scientists plug in CC-PEV and SPAM features to DyFA to decrease their feature dimensions. Accuracy for SPAM has enhanced by 9$\ -\ $15 percent, and CCPEV shows an improvement of 10$\ -\ $13 percent. The results reveal that DyFA outperforms IFAB. However, DyFa has still not been tuned to decrease time complexity.

\section{Image steganalysis using grey wolf optimizer}

Mirjalili \emph{et al} \cite{mirjalili2014GWOproposed} proposed a new evolutionary algorithm based on the concept of grey wolf packs named Grey Wolf Optimizer (GWO). GWO outperforms other evolutionary algorithms in searching for the solution of nonlinear functions in multidimensional space. The GWO algorithm mimics the leadership hierarchy and hunting mechanism of grey wolves. The leadership hierarchy is simulated based on the behavior of different types in pack of wolves, such as the alpha, beta, delta, and omega. Basically, the GWO, inspired by grey wolf hunting procedure, involves three main phases. First, seeking for prey, which is called exploration, encircling prey and lastly, attacking prey, which provides exploitation. This process ensures that scientists identify the global optimum to tackle optimization problems. 

The GWO is a population based algorithm which employs a collective behavior of wolves for seeking the optimal solution. GWO starts with the exploration of search space and exploits gradually to identify the local optimum. GWO provides the most important parameters for adjusting step size by a parameter named "A", and controlling convergence by the exploration and exploitation parameters. The GWO is well-known for its low computational cost. However, it still has some limitations like slow convergence rate and getting stuck in local optima at times. It is obvious that controlling the exploration and exploitation trade-off, defined by "A", plays the main role in GWO.

Pathak \emph{et al} \cite{pathak2019GWO} proposed a new version of GWO for solving steganalysis using feature selection called levy flight-based grey wolf optimization (LFGWO). LFGWO overcomes the limitations of GWO and aims to seek the most prominent features in feature space. The fitness function includes one of the decision tree classifiers, called random forest. The main advantage of LFGWO compared with the relevant literature of evolutionary algorithms is its better convergence precision. They also examined 5 different classifiers - SVM, LDA, RF, KNN and ZeroR - to analyze the performance of image steganalysis over selected features. It is worth mentioning that the LDA here is used as a classifier. LDA also is used as a feature reduction algorithm \cite{ch1_farid}. The LFGWO extracted 84 out of 686 and 89 out of 1000 features from SPAM and AlexNet extracted features, respectively. Although LFGWO obtained better results than IFAB\cite{Farid2014IFAB} and and advanced IFAB, called IFAB-KNN \cite{mohammadi2014IFABKNN}, IFAB and IFAB-KNN extracted less features, i.e. 80 out of 686 for SPAM. The results reveal that the 84 selected features were more relevant than those 80 features, particularly for SPAM.

Veena \emph{et al} \cite{veena2019GRASPBGWO} introduced an optimized method to attack a well-known steganography algorithm, so called Least Significant Bit (LSB). The authors aimed to seek optimal features by the proposed hybrid technique of Greedy Randomized Adaptive Search $\ -\ $ Binary Grey Wolf Optimization (GRASP-BGWO). They succeeded in enhancing classification accuracy of the ensemble logistic regression classifier while shrinking the feature size.

Veena \emph{et al} \cite{veena2019GRASPBGWO} introduced five different spatial LSB algorithms: LSB Replacement (LSBR), LSB Matching (LSBM), LSBM Revisited (LSBMR), Two bit LSBR (LSBR2) and Modulo 5 LSBR (LSBRmod5). The authors applied GRASP-BGWO and observed that the detection process is highly dependent on three important properties: training algorithms, payloads and features. According to their research study, GRASP-BGWO excelled all existing works like SRM \cite{fridrich2012SRM} (Spatial Rich Mode), PSRM (Projected SRM) \cite{holub2013PSRM} and SPAM, even in low volume payload per pixel. GRASP-BGWO improved performance up to 13 percent. In addition to the given GRASP-BGWO and it's performance on LSBR, Shojae Chae-ikar and Ashmadi \cite{chaeikar2019ensemblSW} presented a novel ensemble Similarity weight (SW)-based image steganalysis. The ensemble SW steganalysis comprises three main steps. First is SW analysis, the second step is to adopt SVM classifier, and the third step makes a decision of stego vs cover. In SW analysis, the researchers compute the pixel and channel similarity weights of the given object and generate PSW and CSW datasets. Then, they compare the datasets with their corresponding reference profiles. The last step takes the generated datasets from the second step to make the final decision on whether the image is stego or cover.
 
 \section{Image steganalysis using Pine Growth Optimization}
Pine Growth optimizer (PGO) is inspired by pine tree growth pattern. Pine is a type of tree that creates two branches at each stage (i.e. time period or height) while growing. Initially, it has 0 branches (k=0, k denotes the number of branches on the tree). After that, the tree grows to a particular height, and it creates branches on two opposite sides (first set of branches as k=1). A new pair of branches are produced at every stage accordingly (k=2,3,...). PGO performs the feature selection process based on the pine's growth pattern. It is noteworthy that the pine tree growth may be altered based on the problem requirements. \cite{Guttikonda2019}.
  
The main advantages of PGO are as follows \cite{Guttikonda2019}: 1) It helps any classifier distinguish stego images from cover images. 2) Selecting the most relevant features using PGO helps any classifier improve steganlaysis performance. 3) PGO selects the relevant subsets of features that are sensitive enough to work with the steganographic scheme. 4) Based on this study, PGO also acts as a solution for CoD.
 
\section{Generative dataset for steganalysis}
In the aforementioned research studies, authors mostly took advantage of well-known raw image datasets, such as Breaking Out Steganography System (BOSS) (BOSSbase 1.01), 1000 Pictures, and Photobucket in order to extract features for a goal of classification, specifically image steganalysis.

\noindent $\diamond$ \textbf{SPAM} Researchers have adopted the most common feature extractors. One of them is proposed in a research study presented by Pevny \cite{pevny2010steganalysisSPAM} for Subtractive Pixel Adjacency Matrix (SPAM) feature extraction from the spatial domain of the digital images. SPAM has two clusters of features: one group derived from markov-horizontal and vertical and another extracted from markov-major and minor diagonal features. Merging these two groups brings the total number of SPAM features to 686.
 
\noindent $\diamond$ \textbf{CC-PEV} Researchers also propose CC-PEV features \cite{pevny2007mergingCCPEV}, from transforming the domain of the digital images. It worth mentioning that CC-PEV involves 548 feature vectors which provides 274 features from the original given image and the second half from the calibrated respective image\cite{kodovsky2009calibrationCCPEV}.

 \begin{figure}[ht]
    \centering
  \includegraphics[height=1.3in]{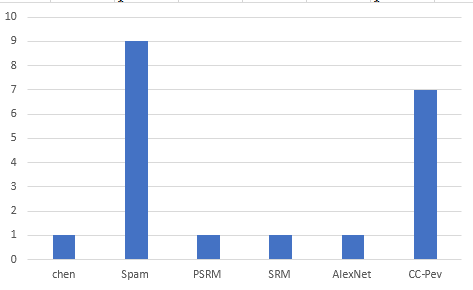}
    \caption{Frequency usage of Feature extractors}
    \label{fig:FEBar}
\end{figure}

 \noindent $\diamond$ \textbf{AlexNet}
The third feature extractor found in literature is AlexNet \cite{krizhevsky2012Alexnet}, which is a deep convolutional neural network that extracts the features directly from images. \par
 \noindent $\diamond$ \textbf{SRM and PSRM}
There are still other state-of-the-art feature extractors that researchers take into account: Spatial Rich Model (SRM) \cite{fridrich2012SRM}, which provides 34,671 features; projected SRM (PSRM), which extracts 12,870 features \cite{holub2013PSRM}; and CHEN\cite{chen2008jpeg}, which is the first common feature extractor that generates 486 features.

\section{Algorithms efficiency} 
The learning process to train a model using training data plays a pivotal role in machine learning. The trained model helps researchers predict test data with high precision. When the number of features dimensions are too high, the learning process takes a long time or the trained model will perform poorly due to the CoD. Due to mentioned obstacles, data analytics phases take longer time and do not match well with visual computing solutions. To do real-time processing, a large number of studies have been explored in this paper to provide a trend in the world of image processing. All the studies state, and it goes without saying, that when higher number of resources and processors are available, the chance of computing in real-time increases \cite{Farid2014IFAB}. While reviewing the above-mentioned studies, we found a direct relationship between the efficiency of the algorithms for data analytics on visual computing and the richness of systems' hardware.

\section{Further optimization and efficient data analytic trends}
All scientific studies discussed in this paper rely on an optimization procedure implemented over evolutionary algorithms with imaging applications. Mohammadi and Saniee abadeh mostly focused on EAs and provided a comprehensive analysis of image steganalysis in \cite{Farid2012}.  Mohammadi \emph{et al} \cite{ch1_farid} also investigated a large number of research studies to tackle high dimensional data using evolutionary algorithms and their applications on real-world problems. Further, Farahani \emph{et al} \cite{farahani2020brief} aimed to tackle unsupervised domain adaptation and worked on groups of the domain adaptation from different prospectus.  Scientists aimed to collect the most important and significant research studies with a high impact on large-scale engineering and science problems \cite{shenavarmasouleh2020drdr}. 
Aside from EAs, many researchers proposed feature selection algorithms based on machine learning and deep learning to tackle high dimensional data in various fields and all of them reported state-of-the-art results from them \cite{zhang2019cost}  \cite{brezovcnik2018swarm} \cite{asali2020deepmsrf} \cite{mohammadi2019parameter} \cite{soans2020sa}.

Additionally, Kumar \emph{et al} \cite{kumar2020curse} presented a randomised smoothing technique to deal with CoD. There is also a wide range of distributed optimization algorithms for complex engineering problems, e.g.image captioning using\cite{amirian2019image}, distributed methods for smart cities applications \cite{amini2019distributed} and decentralized artificial intelligence for energy systems \cite{imteaj2019leveraging}. 
These methods lend themselves as efficient computing tools to deal with large-scale decision making problems in a decentralized/distributed fashion. 

\begin{figure}[ht]
    \centering
  \includegraphics[height=2in]{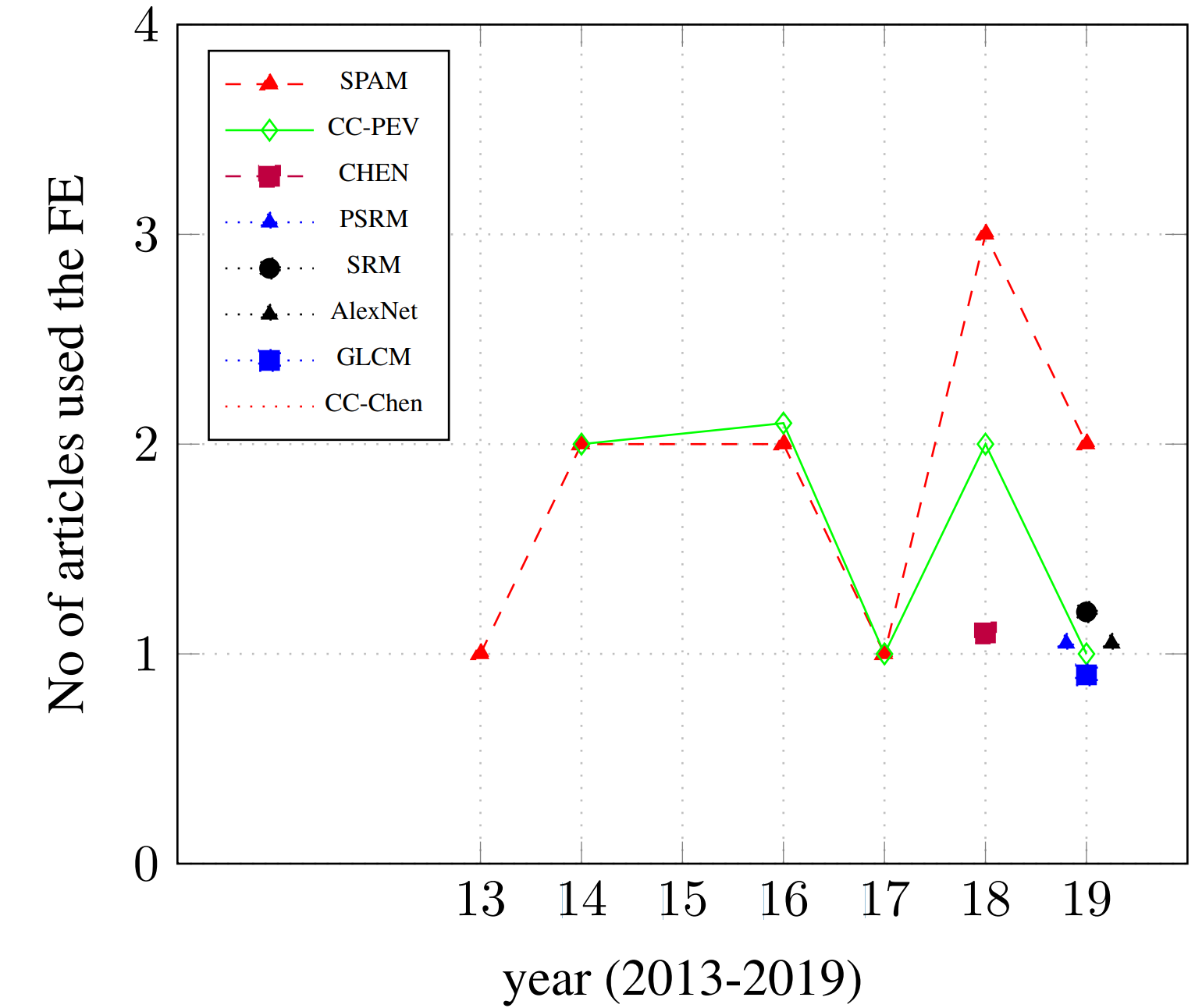}
    \caption{Feature extractors usage within 2013-2019 }
    \label{fig:fig4}
\end{figure}

\section{Discussion and conclusion}
In recent years, there has been significant development in image steganalysis, and researchers are proposing a universal steganalysis to attack all discovered and upcoming steganography algorithms successfully. However, image steganalysis algorithms always suffer from shortcomings; according to the papers considered in this study,  the common problem with state of the art image steganalysis algorithms is the curse of dimensionality. This problem may cause the algorithms to fail to make a proper model and distinguish stego images from cover ones. Not only that, but this problem will also cause the time complexity of the learning process to increase.

It is worth mentioning that the majority of researchers use SPAM and CC-PEV as their feature extractors according to figure \ref{fig:FEBar} and figure \ref{fig:fig4}. Although SPAM is the most commonly used feature extractor, other feature extractors, like SRM and PSRM, have shown good results as well. 

\begin{figure}[ht]
    \centering
  \includegraphics[height=2in]{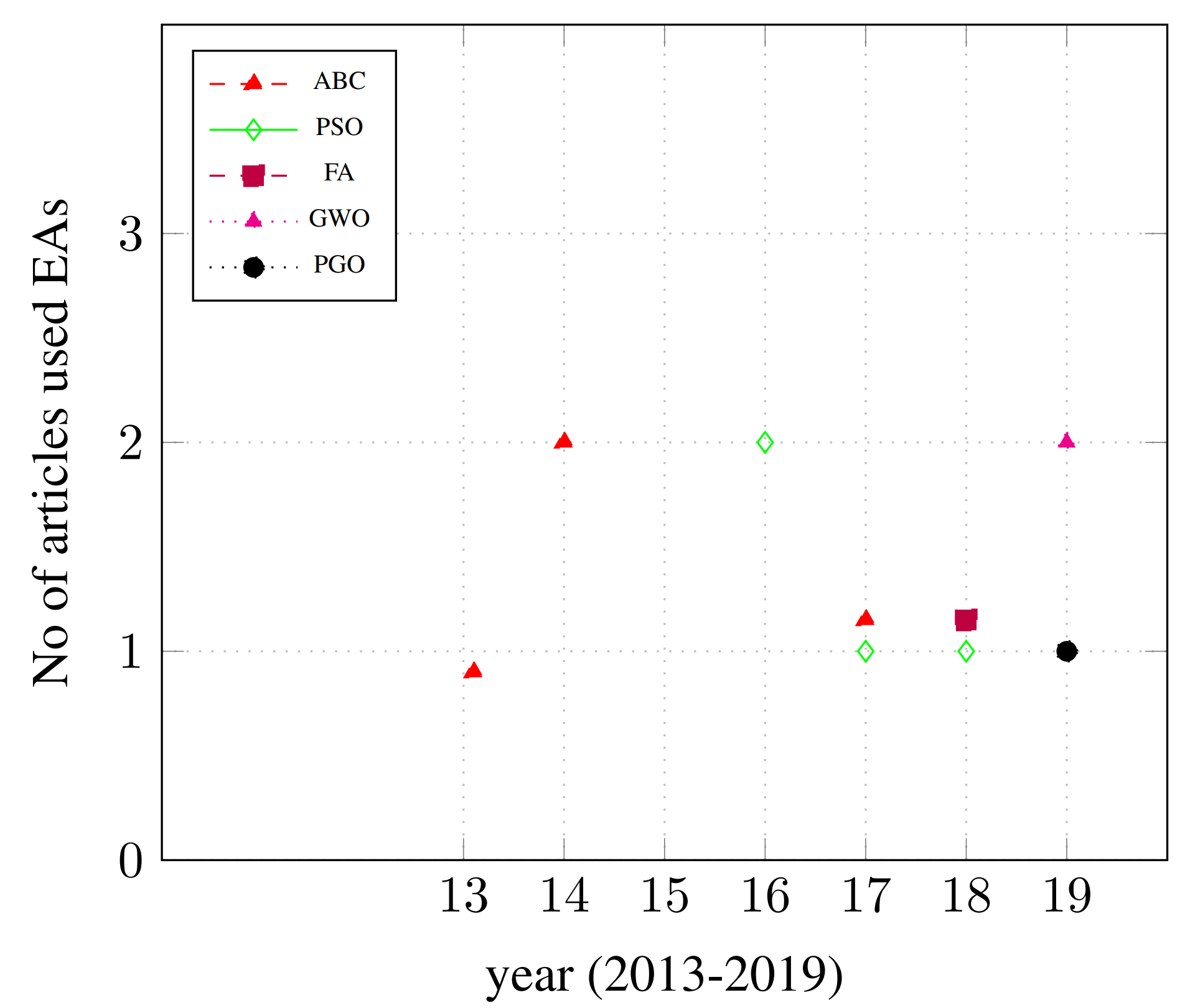}
    \caption{Evolutionary algorithms usage within last decade for dimension reduction}
    \label{fig:fig5}
\end{figure}

Evolutionary algorithms are the most common algorithms associated with high-dimensional data and complex engineering and science problems. In this study, we explored those algorithms that were adopted to solve the shortcomings of image steganalysis with respect to the curse of dimensionality. Five evolutionary algorithms succeeded in alleviating this problem while improving the performance of image steganalysis: Artificial Bee Colony, Particle Swarm Optimization, Firefly Algorithms, Grey Wolf Optimizer and Pine Tree Optimization. Figure \ref{fig:fig5} shows the use of EAs for feature selection in image steganalysis within years 2013-2019.


EAs are used for optimization purposes to ensure real-time image processing capabilities by tackling issues such as CoD. To that end, they are mostly adopted for feature selection. Table \ref{tab:comparisonResearches} illustrates that the majority of EAs used for steganalysis work for different types of feature selection. The table shows that most research studies of EAs use wrapper-based feature selection, followed by filter-based feature selection. 

Evolutionary algorithms have been adopted for image steganalysis since 2013 \cite{Farid2013Conference}. Researchers started using ABC, a powerful optimization tool tailored for image processing. A number of research studies with innovative solutions have been proposed since then to improve the aforementioned traditional evolutionary algorithms. However, they also had their own shortcomings causing the algorithm to fail in properly converging to find the global optimum. There are reasons for failure such as: sensitivity to initial and noisy conditions, being greedy, having biased assumptions, lacking fine-tuned parameters, and being undeterministic. Even though the majority of EAs suffer from high computational costs, they are quite robust in finding the best solution.

The goal of this paper was to provide a comprehensive overview of evolutionary-based image processing for researchers who are currently working or will work on real-time image processing, particularly in image steganalysis. In this research study, we reveal that a limited number of evolutionary algorithms have been utilized for image steganalysis thus far. However, a large number of evolutionary algorithms have yet to be considered to overcome real-time image processing challenges. Therefore, future works should explore other evolutionary algorithms and try to extend them in order to make them suitable for real-time image processing.

\bibliographystyle{unsrt}
\bibliography{bib.bib}
\vspace{10pt}
\end{document}